\documentclass[runningheads]{llncs}
\usepackage{natbib}
\usepackage{graphicx,amsmath}
\usepackage[T1]{fontenc}
\usepackage[misc]{ifsym}

\graphicspath{{./}{./figs/}}

\usepackage{nameref,hyperref}
\usepackage[table]{xcolor}
\hypersetup{
    colorlinks = true,                    
    citecolor = {blue},
    linkcolor = {teal},
           }

\makeatletter
\def\fnum@table{Table~\thetable}
\renewcommand{\thetable}{\arabic{table}}
\makeatother

\begin{document}
%
\title{Small transformer architectures for task switching}
%
%
\author{Claudius Gros$^{\,\scalebox{0.75}{\mbox{\Letter}}}$}
%
\authorrunning{C.~Gros}
%
\institute{Institute for Theoretical Physics, 
Goethe University Frankfurt, Frankfurt am Main, Germany}
%
\maketitle              
\begin{abstract}        
The rapid progress seen in terms of 
large-scale generative AI is largely based 
on the attention mechanism. It is conversely 
non-trivial to conceive
small-scale applications for which 
attention-based architectures outperform
traditional approaches, such as multi-layer 
perceptrons or recurrent networks. We examine 
this problem in the context of `task switching'. 
In this framework models work on ongoing token 
sequences with the current task being determined 
by stochastically interspersed control tokens.
We show that standard transformers cannot solve
a basic task switching reference model based on 
finite domain arithmetics which contains subtasks 
dedicated to 
increment\,/\,addition\,/\,reverse copy\,/\,context
(IARC). We show that transformers, 
long short-term memory recurrent networks (LSTM),
and plain multi-layer perceptrons (MLPs) achieve 
similar, but only modest prediction accuracies. 
We enlarge our comparative
study by including an extension of the standard 
transformer architecture to its non-translational 
invariant counterpart, the cisformer, and an 
alternative attention mechanism, extensive attention. 
A combination of the latter is found to be the only 
model able to achieve considerable performance levels,
of around 95\%.
Our results indicate that the workings of 
attention can be understood better, and even
improved, when comparing qualitatively
different formulations in task-switching
settings. 

\end{abstract}
 
\section{Introduction}

An undisputed advantage of the transformer
architecture is that memory requirements scale 
only with layer depths, but not with context 
length \citep{vaswani2017attention}. 
A feature shared with fully-connected multi-layer 
perceptrons and recurrent networks is that
compute scales quadratically with context length.
It remains an open question whether the success 
of transformers is due to particular properties 
of the underlying attention mechanism, or a 
consequence of the resulting improved size scaling.
Alternative models with favorable scaling 
may be just as good in the later case
\citep{gu2023mamba}. This question received
further urgency by the recent observation
that MLPs learn in-context on par with
transformers when given the same compute budget 
\citep{tong2024mlps}. It is hence important 
to study to which extent transformers excel 
or fail for small-sized applications, namely in
a regime where scaling is not yet relevant.

Here we work below the scaling regime
\citep{kaplan2020scaling,neumann2022scaling},
focusing in particular on setups for which 
transformers and classical models have similar 
numbers of adjustable parameters. We evaluated 
to which extent several standard and new models 
are able to switch the task to be performed upon 
the appearance of suitable control tokens, with
individual tasks working on sequences of encoded 
numbers. We develop a basic version the IARC task,
as explained in Sect.~\ref{sect_IARC}. The 
resulting evaluation protocol can be varied and/or 
extended by selecting appropriate subtasks, making 
it a versatile tool for comparative performance 
tests. Task switching protocols are relevant
for real-world applications, e.g., for steering 
robots by switching between motor 
primitives \citep{saveriano2023dynamic},
in general however with larger networks.

Our results support the notions that
small transformers are not generically
better than MLPs or recurrent LSTM networks.
In our comparative study we include two
novel extensions, the cisformer and expressive
attention \citep{gros2024reorganizing}. The
first is a straightforward generalization of
the standard transformer to the case that
every position along the context dimension
has its own set of adaptable parameters. For
the second, expressive attention, a rational
expression is used for constructing the 
attention matrix out of the dot products 
between keys and queries, instead of the 
usual softmax operation. We find that the
combination of cisformer with expressive
attention is the only model able to achieve
substantial performance level on the basic
IARC task.

\begin{figure}[t]
\centering
\includegraphics[width=0.48\textwidth]{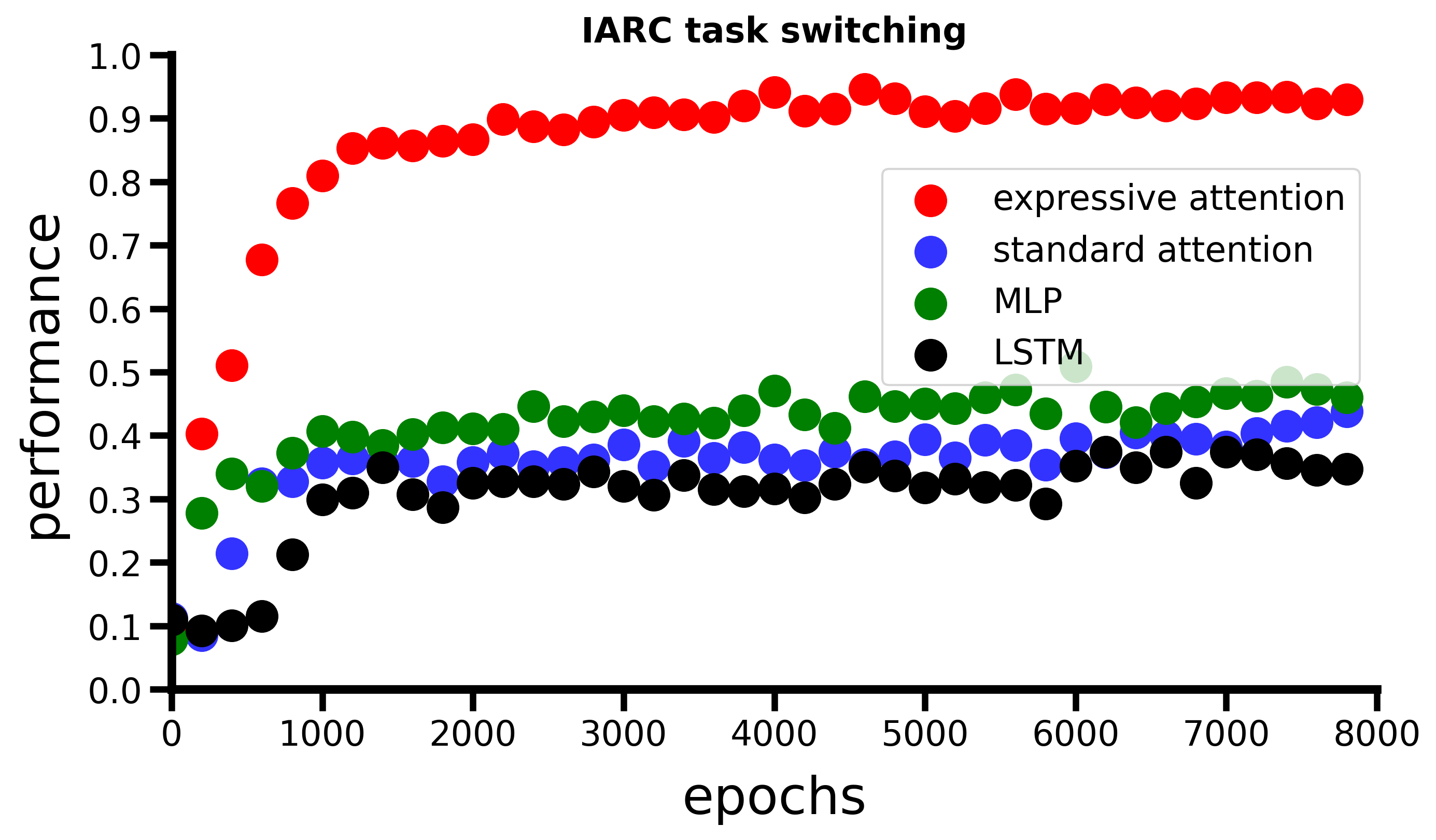}
\includegraphics[width=0.48\textwidth]{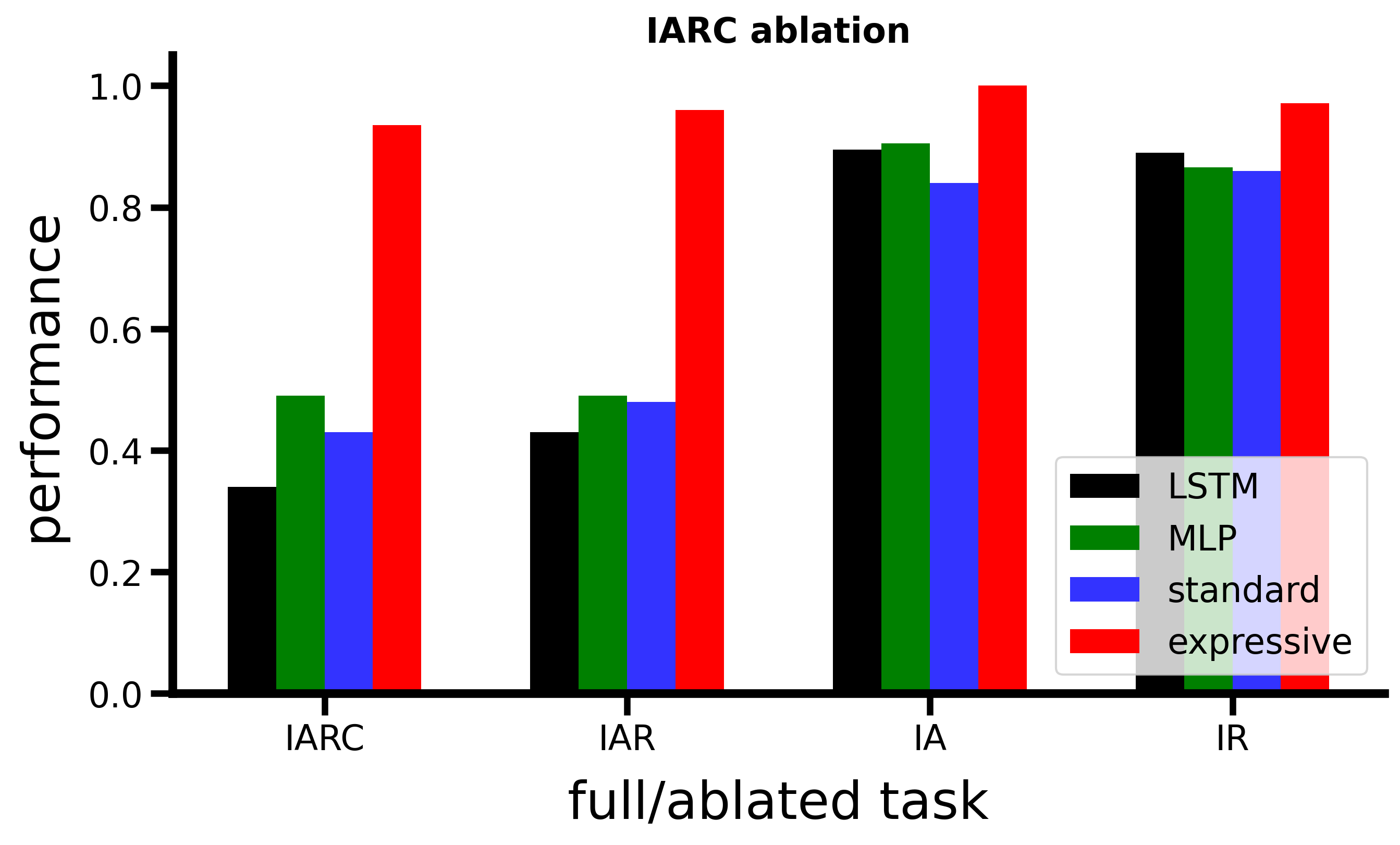}
\caption{Evaluation of the IARC task switching 
framework defined in Sect.~\ref{sect_IARC}.
Shown are results for a LSTM recurrent 
network (black), a MLP (green), and two versions
of cisformers (see Sect.\,\ref{sect_cisformer}),
namely with standard attention (blue), and
with expressive attention (red). 
The equivalent results for classical transformers 
are listed in Table\,\ref{table_standard}.
{\bf Left:} As a function of training epochs, the 
prediction accuracy (performance). 
{\bf Right:} Testing with various 
combinations of the fundamental 
tasks, including
(I), incrementing the current input, 
(A), adding the last two inputs and 
(R), reverse copy. Recursive context 
dependency is encoded by (C). 
}
\label{fig_all}
\end{figure}

\section{Related Work}
\label{sect_related_work}

The task-switching protocol used here is
a specification of multi task learning
\citep{zhang2021survey,chen2024multi}, 
in the form of ever ongoing sequences 
of concatenated tasks. Task switching 
is in particular important for 
reset-free robotic applications 
\citep{gupta2021reset}, e.g., in the
context of embodied robotics 
\citep{kumar2024robohive,sandor2018kick}, 
or for large language models \citep{knight2024multi}. 
Related to our approach are testing procedures 
involving synthetic reasoning tasks 
\citep{zhang2022unveiling}, which have 
been applied to small models in the 
form of in-context and global bigrams 
\citep{bietti2024birth}. Another example
are toy models of superposition, which
can be used to study polysemanticity 
\citep{elhage2022toy}.
It is likewise important to find out
which formal languages transformers
can express \citep{strobl2024formal,deletangneural}.
Equivalently the question arises to
which extend large-model scaling
\citep{kaplan2020scaling,hoffmann2022training},
and variants thereof
\citep{naveed2023comprehensive,shen2024scaling},
is retained when models are small
\citep{ivgi2022scaling}. Equivalent questions
arise in the context of reinforcement learning
\citep{neumann2022scaling,neumann2024alphazero}.

\section{IARC Task Switching Framework}
\label{sect_IARC}

A considerable number of benchmark tasks
for the evaluation of transformer variants 
have been developed \citep{liu2024lost},
however nearly exclusively for large model 
sizes \citep{tay2020long}. The benchmark
task introduced here is meant in contrast
for the evaluation of small models, typically
with a few million parameters or less.

The vocabulary consists of a finite set of 
$N$ numbers, plus a limited number of control
tokens, denoted here (I), (A), (R), and (C),
where I\,/\,A\,/\,R\,/\,C stand respectively for
increment\,/\,addition\,/\,reverse copy\,/\,context.
Control tokens are stochastically interspersed into 
the sequence of symbols, $S_t=\{x_{t'} | t'\le t\}$. 
The task is to predict the next symbol, $x_{t+1}$, 
but not the occurrence of future control tokens. The 
last control token, or the history of previous control 
tokens, determines the dependence of $x_{t+1}$ on 
$S_t$. One has
\begin{equation}
x_{t+1}\big|_I=(x_{t}+1)\%N,
\quad\qquad
x_{t+1}\big|_A=(x_{t}+x_{t-1})\%N
\label{def_I_A}
\end{equation}
for increment (I) and addition (A). 
For $N=10$, an example would be
\begin{equation}
\stackrel{\phantom{\rm A}}{|}   2
\stackrel{\phantom{\rm A}}{|}   3
\stackrel{\phantom{\rm A}}{|}   4
\stackrel{{\color{blue}\rm A}}{|}  7
\stackrel{\phantom{\rm A}}{|}   1
\stackrel{\phantom{\rm A}}{|}   8
\stackrel{\phantom{\rm A}}{|}   9
\stackrel{\phantom{\rm A}}{|}   7
\stackrel{{\color{blue}\rm I}}{|}  8
\stackrel{\phantom{\rm A}}{|}   9
\stackrel{\phantom{\rm A}}{|}   0
\stackrel{\phantom{\rm A}}{|}   1 
\label{example_A_I}
\end{equation}
where time flows from left to right.
Two control tokens appear in this example,
as indicated by the respective superscripts.
Including reverse copy (R), an example is
\begin{equation}
\stackrel{\phantom{\rm A}}{|}   2
\stackrel{\phantom{\rm A}}{|}   3
\stackrel{\phantom{\rm A}}{|}   4
\stackrel{{\color{blue}\rm A}}{|}  7
\stackrel{\phantom{\rm A}}{|}   1
\stackrel{\phantom{\rm A}}{|}   8
\stackrel{{\color{blue}\rm R}}{|}  8
\stackrel{\phantom{\rm A}}{|}   1
\stackrel{\phantom{\rm A}}{|}   7
\stackrel{\phantom{\rm A}}{|}   4
\stackrel{\phantom{\rm A}}{|}   3
\stackrel{{\color{blue}\rm R}}{|}  3
\stackrel{\phantom{\rm A}}{|}   4
\stackrel{\phantom{\rm A}}{|}   7
\label{example_A_R_R}
\end{equation}
Tasks remain when (I/A) tokens are 
followed by (I/A) tokens, with the
reverse copy process being restarted
by subsequent (R) tokens. The action of
the context token (C) depends on the
current task, consecutively increasing 
the increment by one when the task is to
increment:
\begin{equation}
\stackrel{\phantom{\rm A}}{|}   1
\stackrel{{\color{blue}\rm I}}{|}  2
\stackrel{\phantom{\rm A}}{|}   3
\stackrel{\phantom{\rm A}}{|}   4
\stackrel{{\color{blue}\rm C}}{|}  6
\stackrel{\phantom{\rm A}}{|}   8
\stackrel{\phantom{\rm A}}{|}   0
\stackrel{\phantom{\rm A}}{|}   2
\stackrel{\phantom{\rm A}}{|}   4
\stackrel{{\color{blue}\rm C}}{|}  7
\stackrel{\phantom{\rm A}}{|}   0
\stackrel{\phantom{\rm A}}{|}   3
\label{example_I_C_C}
\end{equation}
When the current task is (A/R), the
context token (C) just acts as an
additional (A/R) token, as illustrated
in (\ref{example_A_R_R}) for the case
of two consecutive (R) tokens. As a basic
protocol regulating the frequency of task
switching we use a $6\pm3$ setup, which 
means that the distance between subsequent 
control tokens is drawn from a flat 
distribution out of $[3,9]$. The four 
control tokens, (A/I/R/C), appear with
equal probabilities.

\subsection{Embedding auxiliary tokens via control tapes}
\label{sect_tape}

In a basic encoding scheme, control tokens
regulating task switching would be interspersed
into the input context. For convenience we 
use here a straightforward alternative 
encoding scheme, namely to tape auxiliary 
tokens via a suitable enlarged embedding 
dimension for token activities $x$:
\begin{equation}
x = (x_{\rm sym} | x_{\rm tape})\,,
\label{def_tape}
\end{equation}
where $x_{\rm sym}$ denote symbols activities
with $x_{\rm tape}$ encoding control tokens.
For one-hot encoding, f.i., symbol activities 
are vectors of dimension $N$, with the tape component 
$x_{\rm tape}$ having dimension $N_{\rm control}$, 
where $N_{\rm control}$ is the number of
distinct control tokens. For IARC one has
$N_{\rm control}=4$. Overall, the embedding
dimension is $d=N+N_{\rm control}$.
Of course, control tapes can be used for
any common embedding scheme, besides 
one-hot encoding. Tape activities $x_{\rm tape}$ 
are set to zero when no control token is present.
As an example, the sequence (\ref{example_I_C_C})
is equivalent to
\begin{equation}
\stackrel{{\bf\color{red} I}}{1}\,
\stackrel{\phantom{\rm\color{blue} C}}{2}\,
\stackrel{\phantom{\rm\color{blue} C}}{3}\,
\stackrel{{\bf\color{red} C}}{4}\,
\stackrel{\phantom{\rm\color{blue} C}}{6}\,
\stackrel{\phantom{\rm\color{blue} C}}{8}\,
\stackrel{\phantom{\rm\color{blue} C}}{0}\,
\stackrel{\phantom{\rm\color{blue} C}}{2}\,
\stackrel{{\bf\color{red} C}}{4}\,
\stackrel{\phantom{\rm\color{blue} C}}{7}\,
\stackrel{\phantom{\rm\color{blue} C}}{0}\,
\stackrel{\phantom{\rm\color{blue} C}}{3}
\label{example_tape}
\end{equation}
when a control tape (upper part, red) is used.

\begin{table}[b]
\centering
\caption{For standard, translationally invariant 
transformers with 60 layers and a context length 
of 24, the performance achieved when using the 
original attention mechanism 
(DPA, Eq.~(\ref{DPA})), or expressive attention 
(EA, Eq.~(\ref{REA})). Equivalent results for 
cisformers are shown in Fig.\,\ref{fig_all}.
Tasks abbreviations I/A/R/C stand for 
increment\,/\,addition\,/\,reverse copy\,/\,context.
}
\label{table_standard}
{\setlength\arrayrulewidth{1pt}
\def\arraystretch{1.2}
\begin{tabular}{r|cccc}
& \ \ IARC\ \  & \ \ IAR\ \  & \ \ IA\ \  & \ \ IR \ \ \\ \hline
DPA \ & 0.45 & 0.48 & 0.80 & 0.84 \\
 EA \ & 0.58 & 0.70 & 0.99 & 0.92 
\end{tabular}}
\end{table}

\section{Transformer architectures}

The maximal distance between control tokens
is 9 when a $6\pm3$ setup for token frequencies
is used, as done here. The reverse copy task 
needs consequently a context length $N_{\rm con}$ 
of at least $2\cdot9=18$. From the perspective of 
the model, the appearance of a control token is a 
non-predictable stochastic event. This
implies that the state of the input sequence
further in the past does not convey information 
directly relevant for the execution of the current
task. As a default we use $N_{\rm con}=24$, 
which is larger than 18, but otherwise 
comparatively small.

\subsection{Standard, translational invariant transformers}
\label{sect_classical_transformer}

As originally defined \citep{vaswani2017attention},
transformers are translationally invariant, 
encoded per layer by five matrices 
(query/key/value and two weight matrices for 
the FFN sublayer), together parameters linear
in the embedding dimension $d$. These five 
matrices are broadcasted along the 
context dimension. Model size per layer 
is $N_{\rm layer} = 11\,d^2+O(d)$ when an 
expansion factor of four is used for the 
hidden FFN layer.

For the IARC task with $N=16$ and one-hot
encoding for both symbols and tape, as used
here, the embedding dimension is $d=16+4=20$,
which leads to about $N_{\rm layer}=4000$ 
adaptable parameters per layer, a very small number.
Running simulations with 60 layers we found 
performances of around $45\%$ for the full IARC
framework, as shown in Table.\,\ref{table_standard},
and somewhat more for ablated settings. The
values achieved are non-trivial, being above 
the baseline for random prediction, $1/16=0.0625$, 
but otherwise too small for putative real-world
applications. A possible reason for the
observed poor overall performance could be
the limited number of available adaptable 
parameters, namely $60\cdot4000=240\,\mathrm{K}$.
In order to enlarge model sizes, we considered in 
addition an extension of the basic transformer 
architecture, the `cisformer', which breaks 
translational invariance.

\subsection{Cisformer: position-wise matrices}
\label{sect_cisformer}

One could argue, as done above, that translational 
invariance along the context dimension may not be 
a necessary requirement for task switching frameworks.
Motivated by this consideration we study
the performance of architectures that are
identical to standard transformers in 
all aspects but one: tensors containing 
adaptable parameters are not broadcasted 
along the context dimension, instead they are
independent entities for every position. This
means that positions come with their own
set of adaptable parameters. Model size per
layer is now $N_{\rm con}(11\,d^2+4\,d)$,
which is expressly linear in context
length $N_{\rm con}$. This model type, 
the cisformer, is clearly not suitable 
for applications with large embedding 
dimensions $d$ and context lengths. For 
compact applications like the IARC framework, 
cisformers are however a valid alternative.

\subsection{Expressive attention}
\label{sect_expressive}

The attention mechanism is based on the scalar 
product $z_{ij}=\mathbf{Q}_i\cdot\mathbf{K}_j$
between queries and keys, where the indices
denote the respective positions along 
context dimension. These dot products are 
transformed into row-wise normalized probability 
distributions $a_{ij} = A(z_{ij})$, the attention 
matrix, with $a_{ij}\ge0$ and $\sum_ja_{ij}=1$.
Classically \citep{vaswani2017attention},
a softmax operation is used for $A(z_{ij})$,
\begin{equation}
A(z_{ij}) \sim \exp(\beta z_{ij})
\qquad\quad \mathrm{(DPA)}\,,
\label{DPA}
\end{equation}
modulo a normalization constant, where $\beta$ is 
a formal rescaling factor. Here DPA stands for
`dot-product attention'. Alternatively, a 
rational expression for $A(z_{ij})$ has
been proposed \citep{gros2024reorganizing},
\begin{equation}
A(z_{ij}) \sim \frac{z_{ij}^2}{1+z_{ij}^2}
\qquad\quad \mathrm{(EA)}\,,
\label{REA}
\end{equation}
denoted `expressive attention' (EA). A
key difference between these two formulations
regards attention space geometry. Attention
weights are small when queries and keys
are anti-parallel/orthogonal for DPA/EA.
Given that the number of orthogonal directions
is substantially larger in the space of
attention heads than the number of 
anti-parallel directions (there is just
a single one), it has been argued that
EA enhances attention expressivity
\citep{gros2024reorganizing}. We examine
transformers and cisformers with both
DPA and EA, using causal attention with
four heads and ALiBi positional encoding
\citep{press2021train} in all cases.

\section{Results}

For all models the context length is
$N_c=24$, together with one-hot embedding
and a fixed embedding dimension $d=20$.
We did set $d=N+S$, where $S$ is the number
of control symbols, compare (\ref{def_I_A}).
This implies $N=16$ for IARC and $N=17$ 
for an ablated version like IAR. Cisformers 
have $L=12$ layers, causal self-attention with
four heads and ALiBi positional encoding 
\citep{press2021train}, altogether 1.3M parameters. 
The multi-layer perceptron (MLP) has 
$L=16$ layers with causal connections, resulting 
in 1.9M parameters. The LSTM recurrent net was 
somewhat larger, $L=2$ with a size of 3.7M.
For all models we did additional runs with 
increased numbers of layers, finding only marginal 
improvements.

Comparative results are summarized in Fig.~\ref{fig_all}.
The three basic models, LSTM, MLP and the cisformer
with standard dot-product attention, show roughly 
similar performances, both as a function of training 
compute and when ablating the original IARC model. 
A massive outperformance is seen by cisformers 
based on expressive attention \citep{gros2024reorganizing}.

In Table\,\ref{table_standard} we list the
performance of classical transformers with and
without expressive attention. Overall model
sizes are smaller, namely 240K. Again we find
in part substantial performance boosts when
using expressive attention. With DPA,
classical transformers perform on par with 
MLP and LSTM nets (compare Fig.~\ref{fig_all}).

\section{Discussion}

Model and training parameters were selected
with the aim to obtain a fair comparison between
models. Some details:
\begin{itemize}

\item In order not to overload the presentation,
we concentrated on an exemplary set of parameters,
using a context length of 24, a batch size of 
200, together with 8000 epochs. Learning rate
and momentum were kept constant, respectively
at 0.02 and 0.8. This set of parameters was 
used for all architectures examined. 

\item For a fair comparison, we adapted
the number of layers for the different
models, such that overall model size was
comparable. Concretely, we used 12, respectively
16 layers for the cistransformer and the MLP, 
2 layers for LSTM, and 60 layers for the 
standard transformer.

\item
In addition, we explored a range of different settings.
Smaller learning rates and larger context lengths
affects the results only marginally. The latter
was to be expected, given that control tokens
appear stochastically. Reduced numbers of 
layers affects the performance negatively for all
architectures examined. Relative performance 
differentials remains however akin to those shown 
Fig.~\ref{fig_all}.

\item
The results shown are independent of the
initialization procedure and the seed used,
as a consequence of the comparatively large number 
of training epochs. On the average, a steady 
performance is eventually reached, as illustrated 
in Fig.~\ref{fig_all} (left panel). Performance
can be observed to fluctuate around a mean 
level, with the amplitude being a function
of training parameters (learning rate, momentum).
For a test, we did additional runs with
reduced learning rates, finding decreasing 
performance fluctuations, an expected behavior.

\end{itemize}
Special attention deserves the performance 
boosts observed when switching from a 
softmax-based attention mechanism to 
expressive attention, which is interesting 
in several aspects.
\begin{itemize}
\item Expanding an exponential, one obtains in
leading order a linear term (apart from a constant),
the starting point of linear attention models
\citep{katharopoulos2020transformers,wang2020linformer,wu2022flowformer}.
In contrast, expressive attention is intrinsically 
quadratic in the scalar product between queries and 
keys, making it biquadratic in token activities. 
This observation implies that there is room for 
reformulating the core of the attention mechanism. 

\item Classical and expressive attention correspond
to contrasting attention space geometries
\citep{gros2024reorganizing}. Weights are small
for softmax\,/\,expressive attention when
queries and keys are anti-parallel, respectively
orthogonal. The dimensions of the corresponding
manifolds in the space of attention heads may
hence differ substantially.

\end{itemize}
Finally, we remark on the relevance of task switching
in general, and of the particular task switching framework
examined here, IARC. Switching between tasks is relevant
for applications taking place in dynamic environments,
e.g., as it is the case for autonomous driving and robot 
control. It is important, from this perspective, to examine 
whether a given architecture would excels in task switching.
We propose that the IARC framework is a first, basic
step in this direction. Larger scale applications will
however need substantially more elaborated task
switching frameworks. Let us recall the rational
for the four IARC subtasks:
\begin{itemize}
\item I: Incrementing a number is a unary operation, 
involving only a single context token.
\item A: Adding two numbers is a simple binary operation, 
depending on two context tokens.
\item R: Reverse copy is a memory subtask with 
variable memory depth.
\item C: A subtask introducing recursive context 
dependencies.
\end{itemize}
For the implementation, we found it convenient 
to tape the four control tokens, I\,/\,A\,/\,R\,/\,C.
Overall, IARC is a compressed basic framework that can
be extended along several directions, in particular
for the study of larger models.

\begin{credits}
\subsubsection{\ackname} 
We thank Jonathan Berant, Michael Hahn and Maor Ivgi for discussions.

\subsubsection{\discintname}
The authors have no competing interests to declare that are
relevant to the content of this article.
\end{credits}

\bibliographystyle{splncs04nat}

\end{document}